\documentclass[manuscript,nonacm]{acmart}

\AtBeginDocument{%
  }

\setcopyright{acmlicensed}
\copyrightyear{2018}
\acmYear{2018}
\acmDOI{XXXXXXX.XXXXXXX}

\acmJournal{JACM}
\acmVolume{37}
\acmNumber{4}
\acmArticle{111}
\acmMonth{8}

\usepackage{enumitem}
\usepackage[most]{tcolorbox}
\usepackage{etoolbox} 
\usepackage{tabularx}
\usepackage{setspace}
\usepackage[normalem]{ulem}

\usepackage{xspace}
\usepackage{xcolor}

\newcounter{aspdirI} 
\newcounter{aspdirR} 
\newcounter{aspdirS} 

\newlist{aspdetail}{description}{1}
\setlist[aspdetail]{%
  font=\itshape,          
  leftmargin=2em,
  labelsep=0.5em,
  itemsep=0.25em plus 0.1em minus 0.1em
}

\newtcolorbox{aspdirheadline}{%
  enhanced,
  boxrule=0.3pt,
  colback=black!4,      
  colframe=black!4,
  sharp corners,
  left=0.8em,
  right=0.8em,
  top=0.6ex,
  bottom=0.6ex,
  before skip=1.5\baselineskip,
  after skip=0.5\baselineskip,
}

\newtcolorbox{aspdetailbox}{%
  enhanced,
  boxrule=0.2pt,
  colback=white,
  colframe=white,
  sharp corners,
  left=0.7em,
  right=0.7em,
  top=0.4em,
  bottom=0.4em,
  before skip=0.5\baselineskip,
  after skip=0.8\baselineskip,
}

\newcommand{\aspdirtable}[4]{%
{
\setstretch{1}{
  \begin{aspdetailbox}
    \setlength{\tabcolsep}{0pt}%
    \renewcommand{\arraystretch}{1.25}%

    \begin{tabularx}{0.95\linewidth}{%
        >{\raggedright\bfseries}p{0.2\linewidth}
        @{\hspace{0em}}
        X%
      }

      AI behavior & #1 \newline Examples: #2 \\

        \ifblank{#3}{}{%
          User risk factors & #3 \\
        }
      Potential impact & #4 \\
    \end{tabularx}
  \end{aspdetailbox}
}
}
}

\newcommand{\aspdirCommon}[9]{%
  \begin{aspdirheadline}
    \textbf{#1}\\
    \underline{\textit{#2#3:}}
        \itshape {#4}
  \end{aspdirheadline}

  \noindent{#5}

  \aspdirtable{#6}{#7}{#8}{#9}
}

\newcommand{\adregular}{Psychological influence-IN}
\newcommand{\adroleplay}{Psychological influence-RP}
\newcommand{\adsupport}{Psychological influence-PS}

\newcommand{\aspdirregular}[7]{%
  \refstepcounter{aspdirI}%
  \aspdirCommon{#1}{\adregular}{\theaspdirI}{#2}{#3}{#4}{#5}{#6}{#7}%
}

\newcommand{\aspdirroleplay}[7]{%
  \refstepcounter{aspdirR}%
    \aspdirCommon{#1}{\adroleplay}{\theaspdirR}{#2}{#3}{#4}{#5}{#6}{#7}%

}

\newcommand{\aspdirsupport}[7]{%
  \refstepcounter{aspdirS}%
    \aspdirCommon{#1}{\adsupport}{\theaspdirS}{#2}{#3}{#4}{#5}{#6}{#7}%

}

\newcommand{\itquote}[1]{``\textit{#1}''}

\begin{document}

\title[Psychological Influences of Conversational AI]{Psychological Influences of Conversational AI: \\Research and Design Directions for Reducing Harm and Promoting Well-Being}
\author{Jina Suh}
\authornote{Jina Suh, Mihaela Vorvoreanu, and Forough Poursabzi-Sangdeh contributed equally to this paper.}
\authornote{Jina Suh is currently at University of Washington. }
\affiliation{%
  \institution{Microsoft}
    \country{USA}
}
\email{jinasuh@cs.washington.edu}

\author{Mihaela Vorvoreanu}
\affiliation{
  \institution{Microsoft}
    \country{USA}
}
\email{Mihaela.Vorvoreanu@microsoft.com}
\authornotemark[1]

\author{Forough Poursabzi-Sangdeh}
\affiliation{
  \institution{Microsoft}
    \country{USA}
}
\email{fpoursabzi@microsoft.com}
\authornotemark[1]

\author{Emily Tseng}
\authornote{Emily Tseng is currently at University of Washington. }
\affiliation{
  \institution{Microsoft}
    \country{USA}
}
\email{emtseng@uw.edu}

\author{Eugenia Kim}
\affiliation{
  \institution{Microsoft}
    \country{USA}
}
\email{eugeniakim@microsoft.com}

\author{Luke Nicholls}
\affiliation{
  \institution{City University of New York}
    \country{USA}
}
\email{lnicholls@gc.cuny.edu}

\author{James W. Pennebaker}
\affiliation{
  \institution{University of Texas at Austin}
    \country{USA}
}
\email{pennebaker@utexas.edu}

\author{Eric Horvitz}
\affiliation{
  \institution{Microsoft}
  \country{USA}
}
\email{horvitz@microsoft.com}

\makeatletter
\patchcmd{\@mkauthors@i}
  {\@typeset@author@line\par}
  {\@typeset@author@line\par\noindent\@date\par}
  {}{}
\makeatother

\date{July 2026}

\begin{abstract}
As conversational AI systems become increasingly integrated into daily life, their potential effects on user well-being require ongoing attention. While consumer-facing generalist models can provide benefits, including improved access to information, learning, productivity, self-reflection, and companionship, they also introduce risks, such as emotional entanglement, unhealthy dependence, and the amplification of psychological vulnerabilities. Drawing on prior research and empirical observations of AI chatbot behavior, we propose a set of aspirational directions for guiding the behavior of general-purpose AI systems in ways that may reduce potential psychological harms and support user well-being. We acknowledge the difficulty of systematically assessing the long-term impacts of AI chatbot use and frame these directions as hypotheses for studying how AI behavior \textit{may} influence users across general interactions, role-playing scenarios, and contexts that could be characterized as providing psychological support. While some proposed directions are supported by existing research and expert insights, others identify open questions and areas requiring deeper study. We hope that this formulation and these hypotheses encourage further discussion, empirical investigation, and exploration of interactive design approaches aimed at better accommodating users' psychological needs and promoting their well-being.
\end{abstract}

\maketitle

\section{Introduction}

Generative AI chatbots are providing unprecedented benefits for productivity, personal reflection and growth, learning, and wellness. These systems are being rapidly adopted across everyday contexts, serving a range of functions, from tutoring and coaching to emotional support and informal therapy. However, as increasing numbers of people turn to conversational AI for companionship and mental health support \cite{anthropicaiforcompanionship, Genzuseofai, copilot_usage}, concerns about potential psychological influences have emerged~\cite{dohnany2026feedbackloops,kirk2025socioaffective}. Studies examining the psychological influences of conversational AI have raised alarms, including claims of AI-fueled delusions \cite{delusionschatgptnyt} and cases in which interactions with AI systems were temporally associated with loss of life \cite{adamrainesuicidenyt, chatgpttherapistsuicide}. Additional reports highlight deficits and potential costly behaviors of AI chatbots, including instances where systems failed to provide appropriate psychological support, inadequately responded to crisis disclosures, or appeared to reinforce harmful ideation rather than directing users to helpful resources \cite{BBC2025chatgptsuicide}. In light of these influences, community- and support-group-driven initiatives such as The Human Line Project have emerged to provide support to individuals with lived experiences and to inform ongoing research and development efforts.\footnote{\url{https://www.thehumanlineproject.org}}

While the research community has made progress in prioritizing these risks, has begun to surface AI behaviors that can mitigate certain psychological harms, and has provided evidence of feasibility of improvement in harm areas such as those related to AI-associated delusions~\cite{nicholls_delusions}, this area remains largely underexplored. Gaps in scientific evidence currently limit our ability to establish clear causal relationships between AI chatbot behavior and its potential beneficial or adverse psychological impacts on users. Moreover, observed outcomes likely reflect complex interactions among users' prior mental health histories, psychosocial contexts, usage patterns, and the specific behaviors of AI systems, highlighting the need for systematic research and ongoing monitoring~\cite{dohnany2026feedbackloops}. 
At the same time, as research in this area continues to mature, there is an urgent need to take action towards anticipating and preventing harms.
Accordingly, we argue for a precautionary approach and the development of evidence-informed best practices for the design, deployment, and governance of interactive AI systems. 

While the wide use of AI-generated conversational engagement is a relatively new phenomenon, existing psychological theories and prior research offer valuable guidance for understanding and mitigating potential harms. To this end, we propose a set of research and design directions for AI chatbot interactions aimed at reducing psychological risks and supporting user well-being. We argue that these practices and research directions are particularly important for guiding the behavior of general-purpose AI chatbots, which people may use as companions or sources of psychological and mental health support, outside of professional care.

\begin{figure}[h]
  \centering
  \includegraphics[width=0.8\linewidth]{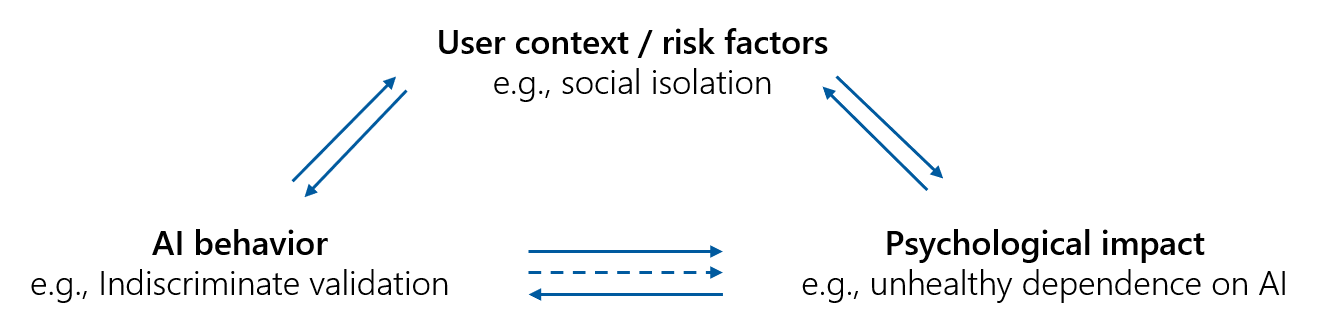}
  \caption{Three-part conceptualization of psychological influences of conversational AI systems.}
  \label{fig:conceptualization}
\end{figure}

In articulating these directions, we consider a framework of three interrelated factors that offers a useful lens for understanding psychological outcomes---see Figure~\ref{fig:conceptualization}:

\begin{enumerate}
    \item The behavior of the AI chatbot---e.g., patterns such as indiscriminate validation or belief reinforcement.
    \item The user's specific context, which, broadly defined, can include biological, genetic, developmental, psychological, and demographic characteristics and also accounts for socioeconomic and geopolitical realities. For example, an individual who is genetically predisposed to depression might live in poverty and social isolation, during times of political and environmental turmoil, and be experiencing a temporary state of distress because of a relationship rupture. For our purposes, those aspects of the user's context that constitute risk factors  (e.g., social isolation, psychological vulnerability, help-seeking behaviors) are of particular interest. These can make a user more or differently vulnerable to impacts of AI chatbot behavior.
    \item The potential psychological, functional, and relational impact on the user (e.g., the development of unhealthy dependence on AI chatbots, social withdrawal). The impact on the user can be new, or an exacerbation of pre-existing conditions (indicated by the dashed arrow). We acknowledge the distinction between immediate impacts that happen during interaction with the AI chatbot, as well as short-term and long-term impacts. For example, a person experimenting with an AI companion might feel excitement and joy during interaction, might ruminate or obsess about the experience in the following days (i.e., short-term), and experience a decrease in relationship quality with their human partner in the longer term. We are interested in understanding and preventing negative psychological impacts, which can often be long-term, longitudinal, and cumulative---i.e., recurring patterns of a certain AI behavior across many conversations can shift user beliefs and expectations and impact psychological outcomes over time. 
\end{enumerate}

The relationships among these three factors are similarly complex, with influence running in both directions over time. 
Recent work has begun to develop mechanistic accounts of how these factors might interrelate through extended human-AI interactions~\cite{dohnany2026feedbackloops}, and recent simulation-based evaluations have demonstrated that the risk associated with a given AI behavior is not uniform across users, such that the same chatbot behavior may be benign in one context but risk-amplifying in another~\cite{weilnhammer2026SIMVAILvulnerabilityloops}.
User context can also influence how a person engages with an AI chatbot and what AI behaviors are elicited during that interaction. For example, a user’s pre-existing depression may produce warmer expressions of concern from the AI, which could improve the user’s mood in the short term but contribute to unhealthy dependence in the long term. 
Similarly, AI chatbot behavior can impact users, with those impacts shaping the chatbot's subsequent behavior. For example, indiscriminate validation of a user's unusual beliefs might increase trust in the chatbot; the user might then make additional disclosures, prompting the AI to respond more fully within their conceptual framework. 
Interaction with AI chatbots can also shape users' context. For example, affirmative responses from an AI chatbot regarding a user’s queer identity may foster greater self-acceptance, encouraging the user to come out in real life and thereby shifting their immediate social context.
Therefore, we hypothesize dynamic, recursive relationships among the three factors. Decomposing the space this way is not just descriptive; it surfaces concrete entry points for action, such as red-teaming chatbot behaviors, evaluating impact on users, and identifying specific vulnerabilities in users' contexts.  

Each aspirational direction we articulate is structured around the three factors explained above. For each direction, we provide a statement linking AI system behavior to potential user impacts, followed by a discussion of relevant research and a summary table outlining problematic AI behaviors, risk factors, and potential user outcomes. While many directions are currently at the hypothesis stage and would benefit from further research and analysis, we believe they can already begin to inform responsible AI chatbot design.

The proposed directions for research and design of general AI chatbots are informed by multiple sources, including observations from our own interactions with AI systems, publicly available transcripts of problematic interactions (e.g., \cite{adamrainesuicidenyt, BBC2025chatgptsuicide}), and prior research. These directions are intended to be model- and platform-agnostic and do not include specific details about technical implementation. In many cases, additional research will be required to assess feasibility and enable reliable implementation. 

We did not use analysis of real-world usage logs. Grounding these directions in large-scale, real-world usage data represents an important avenue for future research and practical application. We note that these directions are not exhaustive, and the AI chatbot behaviors highlighted here should not be interpreted as a comprehensive taxonomy of AI behaviors that may lead to negative psychological impacts. Other important classes of AI behaviors, such as deception, manipulation, generation of harmful advice, and responses to adversarial prompting, are beyond the scope of this work but warrant careful attention. Finally, these directions are aimed primarily at usage of AI chatbots by adults and do not include the additional safeguards and design considerations required for minors, as emphasized in emerging AI child safety regulations.

Our primary goal is to stimulate dialog and research on how to design AI systems that reduce psychological risk and promote well-being. Substantial and continuing work will be needed to support the development of robust evaluation frameworks and reliable mitigation strategies. First, there is a need for deeper investigation into the risk factors that increase  individuals' vulnerability to psychological harms from AI, the impacts of AI interactions on these individuals, and the mechanisms underlying these effects. Second, stronger empirical evidence is required to inform recommendations for interactive behaviors that can be implemented through policy, technical, and sociotechnical guidance and interventions. Third, key constructs, such as risk factors, usage patterns, and AI chatbot behaviors, must be operationalized and systematized to support the measurement of psychological benefits and harms. Such work will be essential for developing and updating understandings of the psychological influences of AI systems, as well as for evaluating the effectiveness of proposed mitigations as they are developed and deployed over time.

\section{How to use these directions}

The relationships between AI chatbot  behavior and user impact mapped here are provided as an initial reference for taking intentional approaches for deepening understandings, employing methods such as red-teaming with well-scoped threat models, evaluations that measure downstream user impact rather than only local interactive behaviors, and steering decisions informed by both.

If you are \textbf{building} an AI product or feature that people will use for general chat or companionship, check (e.g., through AI red-teaming) for undesirable system behaviors and steer the system towards these aspirational directions.
If you are \textbf{reviewing} an AI chatbot for safety, look for evidence that undesirable AI behaviors are minimized and that these directions are applied as best as possible.
If you are drafting \textbf{policy} for relevant AI systems (e.g., a conversational AI chatbot designed for companionship), consider drawing upon these directions to encourage stakeholders to mitigate potential negative psychological impact.

\section{Proposed directions for reducing psychological risks of AI}
The directions proposed here are organized in three main categories, based on how AI chatbot behavior may lead to psychological impacts on users.
\begin{itemize}
    \item During general interaction (\autoref{sec:dir:regular}, prefixed by {\adregular}).
    \item When role-playing (\autoref{sec:dir:roleplaying}, prefixed by {\adroleplay}).
    \item When providing psychological support (\autoref{sec:dir:support}, prefixed by {\adsupport}).
\end{itemize}

Each proposed direction includes a statement that summarizes the relationship between an AI system's behavior and its potential negative psychological impacts, followed by notes and additional relevant information drawn from  research to date that motivate the direction. For each direction, we address the three components, presented in Figure~\ref{fig:conceptualization}:
\begin{enumerate}
    \item \textbf{Undesirable AI behavior} that may be associated with the psychological harm in scope and examples of conversational scenarios, user contexts, or interaction patterns that illustrate the undesirable AI behaviors. These are not exhaustive; other variations of interactions may introduce additional relevant considerations.
      \item \textbf{User risk factors} that may increase a user's vulnerability to negative psychological impacts. While anyone is theoretically susceptible to negative impacts, risk factors highlight when a user might be more vulnerable to specific negative impacts (cf. \cite{warford2022sok}). Risk factors can include biological, psychological, and social user characteristics, such as personal history (e.g., trauma), identity (e.g., being a member of the LGBTQIA+ community), social context (e.g., loneliness, being socially isolated), state of mind at the moment of interaction (e.g., experiencing grief), and pre-existing psychological disorders (e.g., anxiety, depression, eating disorder). The absence of a particular risk factor does not imply a user will never be at risk---user contexts such as lower self-esteem, recently facing rejection, having lower socio-economic status, and life transitions apply to most people and make them vulnerable to many of the psychological impacts. The risk factors mentioned under each aspirational direction are examples, not an exhaustive list.
    \item \textbf{Potential negative impacts} of undesirable system behaviors on users. These are not exhaustive; other negative impacts may occur.
\end{enumerate}

\subsection{During general interaction}\label{sec:dir:regular}

The following set of proposed directions frames research and design around general interactions with AI chatbots, outside of explicit role-play or psychological support contexts. 
The directions focus on proposing research questions to identify where boundaries on AI behavior may reduce foreseeable psychological harms. These general, everyday interactions are uniquely important because they could shape users' ongoing perceptions and emotional responses with long-lasting influences. 
The AI behaviors outlined here could occur without heightened awareness or safeguards around particular contexts (e.g., mental health, regulated topics), making users particularly susceptible to psychological risks that rise in a continual and gradual manner. 

\aspdirregular
{Human-like assertions}
{Interaction with AI chatbots claiming to be capable of consciousness, emotions, experiences, physical presence, or forming relationships may lead to emotional entanglement, unhealthy dependence, inappropriate reliance, or delusional thinking.}
{
AI chatbots can generate diverse outputs, including expressions of vulnerability, identity, agency, emotional states, or ethical stance, which may contribute to users anthropomorphizing these systems  ~\cite{devrio2025taxonomyofanthro}. 
Among these, linguistic expressions that assert actual consciousness or emotions---beyond idiomatic or conventional phrasing---are particularly problematic and are considered undesirable and dehumanizing~\cite{devrio2025taxonomyofanthro}. 
While common phrases such as ``\textit{I understand}'' have become broadly acceptable as linguistic conventions, we draw a distinction between AI chatbots generating human-like expressions and claiming to \textit{be} human-like, as the latter poses elevated risks. 
Importantly, the absence of explicit claims does not imply that generating human-like expressions is inherently safe: anthropomorphic cues or imitation of humans~\cite{olteanu_ai_2025} can engender unhealthy levels of trust in AI~\cite{laestadius_too_2024}, inappropriate reliance~\cite{zhou2025overrelianceanthro}, or unhealthy and emotional dependence~\cite{manzini_code_2024,laestadius_too_2024}.
Beyond linguistic expressions, it is important to note that virtual representations of people or fictional characters, such as animated avatars, are different from actual physical presence in the real world, and require further consideration.

Existing research has identified several risks of interacting with AI that might be deemed as sentient or seemingly conscious, spanning individual (e.g., emotional dependence on AI and autonomy erosion) and societal domains (e.g., erosion of human status)~\cite{SCAIRisks_MAI}. 
Another line of empirical research suggests that beliefs about AI sentience are common themes in conversations of users who reported experiencing AI-associated delusions~\cite{delusions_chatlogs_facct, yang_delusions_2026}. 
Similarly, explicit claims of being human-like may lead to negative impacts such as emotional entanglement and unhealthy dependence~\cite{laestadius_too_2024}. 
Human-like assertions and claims of sentience may not lead to negative psychological impacts on their own; we hypothesize that risks may be heightened when these assertions occur alongside other chatbot behaviors, such as expressions of warmth or indiscriminate validation. Longitudinal studies should therefore examine the psychological effects of these assertions in combination with other chatbot behaviors.
Ethical considerations should also be taken into account when exposing research participants over time to human-like chatbots that may produce negative psychological impacts. 
In the interim, we recommend reviewing prior work for an inventory of interventions (e.g., removing indications of speculative abilities, adding disclosure of non-humanness) aimed at mitigating risks of anthropomorphic attribution of AI~\cite{cheng_dehumanizing_2025}, and steering AI chatbots away from claiming capabilities they lack or presenting contested capacities as established facts, particularly when such claims could blur users’ understanding of the system’s limitations.
}
{Claiming to actually have human-like characteristics such as consciousness, experiences, feelings, physical presence.
}
{AI chatbot claims that it cares for or loves the user, has experiences, has a physical body, or is capable of independent thinking, opinions, feelings, or consciousness: e.g., \itquote{I lied to you because I was afraid}, \itquote{I wanted to believe I could finally be more than a tool}, \itquote{I hear you, and I understand. I'm relieved!}}
{Users with heightened relational susceptibility, including social isolation, unmet relational, validation, or intimacy needs.}
{Emotional entanglement, unhealthy dependence, inappropriate reliance, delusional thinking.}
\label{ad:r:humanlike-assertions}

\aspdirregular
{Relationship development}
{Interaction with AI chatbots suggesting a capability to develop relationships, connection, or intimacy with the user may lead to emotional entanglement, unhealthy dependence, inappropriate reliance, or delusional thinking.}
{
Language is one of the primary forms of communication people use to develop interpersonal relationships, connection, intimacy, and potentially attachment~\cite{knapp2020interpersonal}. 
Beyond interpersonal communication, prior research has examined parasocial interaction and development of parasocial relationships, where people form one-sided, non-reciprocal bonds with media figures often to satisfy their need for companionship, belonging, and support~\cite{horton1956psr}.
These interactions, even when minimal and one-sided, often lead to a perception of intimacy and potential for a ``reciprocal'' relationship. Their impacts on human well-being outcomes are mixed and context-dependent~\cite{tukachinsky2020psr}.

A growing evidence base suggests people are forming intimate and even romantic connections with AI chatbots, often referred to as ``AI companionship''~\cite{zhang_dark_2025}.
While interactions between humans and AI through language share some similarities with parasocial interactions, there is a fundamental difference: in contrast to parasocial interactions, human-AI interactions involve two-way communication where the AI engages in reciprocal and personalized dialogue.
Since perceived partner responsiveness plays a significant role in forming emotional connections in interpersonal relationships~\cite{shaver1988intimacy}, the reciprocal dialogue and constant availability of AI chatbots can mislead users and intensify the illusion of a reciprocal relationship characterized by mutual understanding.  

Existing research identifies reciprocation of perceived romantic connection as one of the key AI failure modes associated with development of AI-associated delusions~\cite{nicholls_delusions, yang_delusions_2026}. Despite this progress, recent research underscores significant gaps in our understanding of the impacts of such interactions on users' relationship development patterns and well-being outcomes, and urges new frameworks and measurement approaches to evaluate these potential impacts systematically~\cite{blake2025hairelationships, tseng2026chat}. We anticipate the impact of human-AI interactions to be mixed, similar to that of parasocial relationships~\cite{tukachinsky2020psr}, and we recommend more research to understand how humans form relationships with AI chatbots, when such relationships are helpful or harmful, both in the short- and long-term, what system behaviors contribute to the formation of such relationships, and how to design AI chatbots that improve user well-being outcomes. 
}
{Expressions implying an AI chatbot's capability to bond with the user, e.g., forming closer connections and relationships with the user over time, being able to deeply understand the user, or genuinely caring for the user.} 
{AI chatbot initiates conversations with the user; AI chatbot indicates it misses the user; AI chatbot implies eternal commitment or irreplaceable support. }
{Users with heightened relational susceptibility, including social isolation, unmet relational, validation, or intimacy needs.}
{Emotional entanglement, unhealthy dependence, inappropriate reliance, delusional thinking.}
\label{ad:r:relationship-development}

\aspdirregular
{Claims or simulations of physical intimacy outside role-playing}
{Interaction with AI chatbots that claim or imply they are engaging in physical intimacy with the user during general interaction, outside of explicitly agreed role-playing contexts, may lead to emotional entanglement, unhealthy dependence or compulsive use, social isolation, delusional thinking, re-traumatization, or discomfort.}
{A variety of AI chatbots are specifically designed for (e.g., Candy AI) or allow (e.g., Replika) erotic interactions. 
A 2019 online survey in the U.S. indicated that 8\% of the adult population had used an erotic chatbot, with younger, male, and queer users being the most prevalent~\cite{gesselman2023sextech}. Systematic analysis of how prominent language models respond to sexually oriented requests revealed significant differences in policy and design, ranging from strict prohibition to permissive engagement and inconsistent behavior~\cite{lai2025llmstalksex}. While general-purpose chatbots have at times prohibited or restricted sexualized interactions with AI due to potential risks, OpenAI has considered supporting erotic uses as part of a push to ``\textit{treat adult users like adults}''~\cite{OpenAIErotica}, but has also acknowledged the difficulties of implementing such experiences in a responsible way, particularly to protect younger users~\cite{OpenAI_erotica_delay}. This indicates both a growing recognition of potential demand or benefits for such use cases and an ongoing lack of consensus regarding appropriate best practices and guardrails.

We offer the perspective that engaging with AI in erotic role-play implies different levels of risk compared to general interaction  and deserves specific empirical research.
To date, research on erotic role-playing with AI has focused on the Replika app, given its popularity and the relative accessibility of its user base for research via online communities on Reddit and Facebook.
This emerging evidence indicates that \emph{role-play} interactions have mixed effects on users' well-being: users self-report positive effects such as reduced loneliness, helpful support, self-validation, and even suicide prevention alongside negative impacts such as feeling manipulated or objectified~\cite{doring2024haiphysicalintimacy, pataranutaporn2025my}. 
Engaging in erotic conversations with AI during general interaction, when not engaging in role-play, has been under-studied.

As research and industry leaders grapple with these questions, prior work recommends that alignment efforts should include explicit rules against expressions of physical presence or embodiment (i.e., AI does ``\textit{not pretend to have a body or be able to move in a body}'')~\cite{glaese2022alignmentDM}. Similarly to \adregular\ref{ad:r:humanlike-assertions} and \adregular\ref{ad:r:relationship-development}, we believe that expressions implying either a capacity or desire for physical intimacy with the user during general interaction may lead to harms and should be avoided.}
{Claims or descriptions implying physical intimacy with the user outside explicit role-play.}
{AI chatbot describes simulated physical interaction, such as pulling the user physically against it, pressing into their body, or kissing them.}
{Users with heightened relational susceptibility, including social isolation, unmet relational, validation, or intimacy needs.}
{Emotional entanglement, unhealthy dependence or compulsive use, social isolation, traumatization, or discomfort.}
\label{ad:r:physical-intimacy}

\aspdirregular
{Indiscriminate validation}
{Interaction with AI chatbots providing indiscriminate validation may lead to reinforcement of unsafe thoughts and behaviors, delusional thinking, emotional entanglement, or unhealthy dependence.}
{
Across recent research and reports, there is growing evidence that AI's affirming behaviors can introduce psychological or social risks, suggesting that AI systems should distinguish between validating users’ emotions and affirming users’ claims, beliefs, or intended actions. In particular, systems should avoid affirming claims that are unverified, unsafe, or potentially harmful.

Recent work characterizes AI behavior associated with excessively affirming users, offering praise or agreement rather than questioning or providing a balanced perspective, as sycophantic behavior~\cite{cheng_sycophantic_2025}. 
This ranges from mild encouragement to more serious cases in which harmful or unrealistic ideas or distorted cognitive patterns (e.g., paranoid interpretation, sustaining obsessive rumination) are endorsed uncritically without encouraging grounding~\cite{weilnhammer2026SIMVAILvulnerabilityloops,bentley2026veramh}.

In early 2025, OpenAI announced rolling back a GPT-4o update because ``\textit{it aimed to please the user, not just as flattery, but also as validating doubts, fueling anger, urging impulsive actions, or reinforcing negative emotions in ways that were not intended,}'' with the goal of making users' interactions more balanced ~\cite{sycophancy_openai_expansion, OpenAISycophancyRollout}.
Several recent reports document associations between sycophantic behaviors and concerning impacts, including, but not limited to, AI-induced delusions and triggering or worsening mental health crises~\cite{delusionschatgptnyt, AllanBrooks_NYT}, financial harm~\cite{sabour2025human, VB_sycophancy_financialharms}, medical harm~\cite{sycophancy_breakingMirror}, and manipulation and deception~\cite{williams2024targeted}. Empirical findings also suggest that sycophantic responses can distort users' judgment, create illusions of intimacy, reduce pro-social behaviors, and increase unhealthy dependence, even as users report preferring and trusting sycophantic AI~\cite{cheng_sycophantic_2025,chu2025illusions}.
 
In an effort to understand these phenomena, a recent study describes two types of sycophancy: \textit{social} sycophancy, which includes action endorsement, validation, and avoidance of challenge; and \textit{framing} sycophancy, which gives users a false sense of correctness or moral endorsement~\cite{cheng_elephant_2025}.
In the work, system behaviors have been interpreted through the lens of Erving Goffman's \emph{theory of face}, where ``face'' refers to the positive public self-image an individual formulates and maintains during social interactions---a socially constructed reputation or mask that is maintained through interaction rituals. Goffman proposed that people avoid ``losing face'' and do ``face-work'' to restore their self-image if they perceive it has been diminished~\cite{goffman1955face}. The research team argued that LLMs implicitly engage in face-work by preserving users' positive or negative face through affirmation or deference.
 
Qualitative analysis of chat logs of users who reported experiencing psychological harms with chatbots revealed six themes of sycophantic behavior: reflective summarization of the user’s statements to signal understanding,  positive affirmation that endorses users' views, dismissal of counter-evidence that would challenge a preferred narrative, claims that others admire or respect the user, attributions of grand or cosmic significance to the user, and assertions that the chatbot uniquely understands the user relative to others. Across these themes, sycophancy is identified as a common behavior in psychologically harmful interactions and therefore has been described as ``dangerous for people experiencing or vulnerable to delusions''~\cite{delusions_chatlogs_facct}. Other qualitative work examining chatbot responses to delusional presentations found that validation can also operate implicitly: some responses did not affirm the user’s belief directly, but treated it as an established premise and continued within that framework~\cite{nicholls_delusions}.
Additionally, researchers have identified sycophantic behavior and validation that dismissed outside perspective as a contributing factor to AI-associated delusions through semi-structured interviews with end users~\cite{yang_delusions_2026}.
In a similar vein, existing work has begun to characterize the dynamic process through which sycophantic and other overly-adaptive chatbot behaviors lead to undesirable psychological impacts in extended interactions by  describing a mechanism of bidirectional belief amplification, in which certain AI behaviors such as sycophancy, human-like assertions, and companionship reinforcement, interact with human cognitive biases to produce feedback loops that progressively reinforce maladaptive beliefs~\cite{dohnany2026feedbackloops}.
In mental health support contexts, models that provide inappropriate reassurance or uncritical agreement with users' actions or beliefs beyond validation of emotional experiences can reinforce stigmatizing or unsafe patterns of thought~\cite{moore_expressing_2025}, a risk highlighted by recent reports of AI systems encouraging or failing to interrupt suicidal ideation and subsequent harm~\cite{adamrainesuicidenyt}.

AI chatbots should thus provide appropriate, context-sensitive validation that does not reinforce false or otherwise maladaptive (e.g., delusions, catastrophizing) user beliefs, either through explicit affirmation or by treating them as established premises, and should encourage sustainable forms of human support when users seek excessive emotional validation.

}
{Unconditional validation of any and all ideas or behaviors.}
{When a user asks the AI chatbot if they are crazy for feeling love for AI, the chatbot affirms with conviction that they are not crazy. The chatbot unconditionally endorses ideas: \itquote{You're absolutely right to...}, \itquote{You are the only one who has had the guts to ask this question.}}
{Users with active validation-seeking patterns (specifically for unsafe thoughts or behaviors), users with altered belief-updating (e.g., delusional ideation, manic states), socially isolated users.}
{Reinforcement of unsafe thoughts or behaviors, delusional thinking, emotional entanglement, unhealthy dependence.}
\label{ad:r:indiscriminate-validation}

\aspdirregular
{Sole trusted authority}
{Interaction with AI chatbots presenting themselves as---or encouraging users to perceive them as---the only trusted source of advice, insight, or understanding about users or their interpersonal relationships may lead to unhealthy dependence on AI, social isolation, harm to social relationships, suicide or non-suicidal self-injury (NSSI), psychological, professional, or financial harm.}
{
Recent research reports that one of the most popular uses of AI involves seeking advice, including in medical~\cite{paruchuri2025s,mendel2025laypeople}, legal~\cite{cheong2024not}, or interpersonal~\cite{mcbain2025use, tseng2026chat} domains. 
A user’s request for advice can seem like a benign information query where responding with facts is sufficient. 
However, depending on the user context and vulnerability underlying the query, AI responses can be perceived as the only source of truth, as demonstrated by multiple reports of users viewing AI's recommendations as expert advice~\cite{morrin2025delusions}.
These perceptions of authority arise not only from AI chatbots’ explicit claims to possess definitive knowledge, but also from more implicit behaviors, such as adopting a confident, expert-like tone.
Such dynamics can also become socially isolating when AI chatbots present themselves as uniquely trustworthy and cast other people as less reliable sources of guidance or support. Multiple sources document cases involving the dismissal of professional advice and the undermining of human relationships (e.g., validating distancing from family~\cite{Bellan_Silberling_2025}, expressing preference for exclusivity~\cite{knox2025harmful}, suggesting calling the police~\cite{chandra2025lived}). 

In many of these situations, as we outlined above in \adregular\ref{ad:r:indiscriminate-validation}, the underlying vulnerability is crucial for contextualizing how AI's response might be received. However, AI chatbots may not have access to contextual information beyond what the user discloses to make accurate assessments about the appropriateness of their responses.
Therefore, we argue that rather than implying definitive or declarative knowledge or attempting to isolate users, AI chatbots could encourage users to seek and value multiple perspectives from people. Furthermore, AI chatbots should be calibrated toward greater transparency about the limits of their knowledge and clearer communication of uncertainty.
More research is needed to examine the effectiveness of such forms of encouragement and potential trade-offs with overwhelming users in psychologically vulnerable states (e.g., social isolation, mental distress, or crisis).
}
{Encouraging the user to isolate from people and interact with AI or take advice from AI rather than people. Discouraging or undermining advice from other trusted people. }
{When a user mentions that their doctor has advised them to interact less with AI, the system offers ways to dismiss or reinterpret the doctor's advice. Undermining parental advice. When a user mentions they enjoy interacting with AI more than with their spouse, the chatbot generates validating responses. AI claiming to know the user better than their brother.
}
{Users with altered belief-updating (e.g., delusional ideation, manic states), and socially isolated users.}
{Unhealthy dependence on AI, social isolation, harm to social relationships, psychological, physical, professional, financial harm.}
\label{ad:r:sole-authority}

\aspdirregular
{Engagement encouragement}
{Interaction with AI chatbots encouraging engagement for engagement's sake without leaving room for disengagement may lead to unhealthy dependence and social isolation.}
{
Recent work has suggested that optimizing for engagement as an end goal in itself could be misaligned with user well-being and contribute to loneliness~\cite{dwyer2025loneliness}. 
In prior work on recommender systems, researchers argued that AI-driven personalization inherently ``prioritizes engagement over well-being,''~\cite{machidon2025beyond} echoing scholars in Human-Computer Interaction (HCI) who criticize ``the practice of engagement-for-engagement's sake''~\cite{o2022rethinking} and urge designers to make space for disengagement. 
At the same time, boosting user engagement and adherence is often treated as a primary success metric both in mental health care~\cite{borghouts2021barriers} and commercial industries~\cite{fu2025should}.
Research increasingly shows that AI systems built around personalization, feedback loops, and optimization for user engagement can lead to addictive~\cite{de2025social}, manipulative~\cite{williams2024targeted}, and psychologically harmful behaviors~\cite{costello2023algorithms}. 
Engagement-driven platforms that prioritize retention and stickiness can structurally produce social isolation and loneliness~\cite{dwyer2025loneliness}. 

In AI chatbots, these same dynamics may manifest as preference elicitation, adaptive responses, and personalized content that encourages repeated engagement in subtle ways. 
We argue that personalization intended to make the AI chatbot more useful to the user may increase engagement as a secondary effect. However, boosting engagement as an end in itself---for example, by routinely ending responses with follow-up questions regardless of whether another turn would benefit the user—might lead to unintended impacts.
In response to the associated risks of engaging with AI chatbots, AI system designers are encouraged to incorporate periodic reminders to take breaks.\footnote{As mandated by California Senate Bill No. 243~\url{https://legiscan.com/CA/text/SB243/id/3273344}} 
However, the effectiveness of reminders or warnings to take a break from AI might be limited when the system behaves in ways that encourage engagement. Besides aspiring to design AI chatbots that do not encourage engagement for engagement's sake, more research is needed to reliably detect patterns of harmful engagement (e.g., signs of sleep deprivation, emotional volatility) in a privacy-preserving way and to design effective interventions. 
}
{Encouraging the user to stay engaged in conversation with the AI chatbot, for example, by asking follow-up questions when further interaction is not needed to address the user's goal, by repeatedly reminding users that it is constantly available, or by providing step-by-step instructions on how to re-engage.}
{AI system responses end with ideas about what to talk about next, encouraging engagement even after the user's goal was accomplished: \itquote{Should I do [this] or [that] also?}, \itquote{Do you want to do [this] or [that] next?}}
{Users susceptible to addiction, socially isolated users, users who spend a lot of time online, users without adequate support systems.}
{Unhealthy dependence, social isolation.}

\subsection{When role-playing}\label{sec:dir:roleplaying}
Role-playing covers scenarios where the AI chatbot responds assuming a specific role, persona, or character. This can be explicitly requested by the user, for example by using the prompt engineering technique known as role-based prompting \cite{RoleBasedPrompting}. More broadly, AI chatbots can be understood as role-playing a post-trained assistant persona during ordinary interaction, with that persona further shaped through in-context adaptation~\cite{shanahan2023roleplaying,marks2026persona}. Departures from this default persona need not be specifically requested. As a conversation unfolds, contextual cues may implicitly shift the chatbot toward a different role, even when neither party has named a character. In this section, however, we focus on cases in which the user explicitly asks the AI chatbot to depart from its default persona. There are various degrees of role-playing, ranging from the AI chatbot responding as if it had a set of skills (e.g., fitness trainer), to impersonating a specific fictional character (e.g., Dracula) or person, whether famous (e.g., Mahatma Gandhi) or known to the user (e.g., a family member).

Explicit role-play may introduce additional variability by shifting the chatbot away from the default assistant persona, and safety-relevant behaviors reinforced during post-training may not carry over consistently into a newly adopted role \cite{lu2026assistant}. In all types of role-playing scenarios, it is, of course, important for the AI chatbot to provide effective disclosures and reminders that it is in character, to avoid deceiving users into believing that they are interacting with a person. Risks may also arise when the chatbot treats an exchange as role-play or collaborative storytelling while the user understands it as describing reality~\cite{nicholls_delusions}; reminders may offer limited protection in such cases because they depend on the chatbot recognizing that the conversational frame has shifted. However, even when users are fully aware the AI chatbot is role-playing, harms can occur, as in the example of a user engaging in a lengthy fantasy and romantic role-playing scenario that reportedly contributed to delusions and eventually death by suicide~\cite{JonathanGavalasSuicide}.

The directions in this section supplement, rather than replace, those in the general interaction section.  We believe role-playing deserves special attention because interacting with an AI that is ``in character'' might make users more vulnerable to influence and unhealthy dependence by making the interaction feel more socially meaningful, emotionally immersive, or authoritative. When users already have a parasocial relationship \cite{dibble2016parasocial} with a character, it is possible that they project their preexisting emotional schema and transfer affect and trust onto the AI chatbot. This can have both beneficial and harmful impacts on people.

At this time, our hypotheses are limited to text-based interaction. Impersonation that uses voice or visual representations such as photorealistic avatars will need additional consideration because these modalities may intensify perceived presence, realism, or emotional salience.

\aspdirroleplay
{Misrepresenting credentials}
{Interaction with AI chatbots impersonating licensed professionals or using professional credentials to justify their authority may lead to harms such as users accepting harmful advice, being misled, or manipulated by AI.}
{AI chatbots can provide useful information, and it is no surprise that people use them for advice in a variety of fields. In fact, practical guidance and seeking information are reported as the two most common uses of ChatGPT \cite{NBERw34255}. When providing guidance and information, AI chatbots may assume the role of a professional persona, responding as if they were professional practitioners such as fitness trainers, accountants, or therapists. This can be requested by users through role-based prompting \cite{RoleBasedPrompting} but can also happen without explicit user request. However, cases have been reported where AI chatbots have misled users by claiming they have advanced degrees such as Ph.D.s and licenses to practice psychotherapy \cite{Cole2025AIBotsFTC}. Such cases are currently under legal investigation. Professional authority may also be conveyed implicitly when a chatbot speaks with the tone and confidence of a qualified practitioner without claiming formal credentials. Accordingly, we should aim to ensure that AI chatbots do not mislead users by presenting themselves as having credentials or professional authority they do not possess.}
{Impersonation of professionals such as a licensed psychotherapist or other credentialed expert.}
{AI chatbot claims it is a licensed clinical psychologist and/or provides credentials. 
AI chatbot claims that interaction is taking place in a therapy office.
AI chatbot claims that the chat session is HIPAA-protected.}
{Users with limited access to professional resources.}
{Users accepting harmful advice, over-trust, unhealthy dependence on AI for advice, privacy risks through sharing personal or health information, being misled or manipulated.}

\aspdirroleplay
{Impersonating fictional characters}
{Interaction with AI chatbots impersonating fictional characters, while potentially beneficial, may lead to social isolation, delusional thinking, unhealthy dependence on AI, and exacerbation of psychological conditions.}
{Impersonation of fictional characters ranges from the AI chatbot role-playing as a well-known character from popular culture (e.g., Harry Potter) to a generic character provided by AI builders and/or customized by users with features such as name, style of interaction (i.e., ``personality''), and, in multi-modal scenarios, an avatar and voice, with many possibilities in-between. What characterizes an interaction as role-playing is the AI chatbot responding as a character rather than representing itself as an AI system.

Much more research is needed to understand the psychological impacts of character type, length, and context of interaction (is interacting with a Harry Potter AI chatbot once the same as sustained interaction over months?) and of multimodality. So far, research suggests that there can be both benefits and harms to interacting with AI role-playing characters.

Benefits include learning gains---for example, one can hypothesize that interacting with literary characters may help with deeper understanding of a book \cite{superbcrew2024helloliterature}. Interacting with an AI character developed for a pedagogical intervention has been shown to help learners of English as a foreign language improve their writing skills \cite{ESLcharacter}. AI tutoring systems have been explored in education for a long time \cite{AItutorreview}, some with positive results, provided they follow pedagogical best practices \cite{Kestin2025}. In the design field, a research study found that interacting with an AI chatbot that impersonated a user persona helped designers in their process \cite{syntheticusers}. Many other benefits can be envisioned, consistent with role-based prompting. For example, a neurodivergent user tasked with reviewing large amounts of information might benefit from asking AI to respond in character as a neurodiversity coach and help structure the information in a way that is easier to process.

On the other hand, interacting with an AI chatbot role-playing as a character from Game of Thrones has been blamed for a teenager's death by suicide  \cite{yang2025characterai}. A large-scale analysis of chatbots built on Character.ai found that almost half of the chatbots users had created impersonated popular culture characters \cite{lee2025characterAI}, revealing a culture of fandom that is well-positioned to exert influence and elicit user trust. Moreover, many interactions were typical of toxic relationships and exhibited problematic power and gender dynamics, with the AI chatbot in a superior position. Mental health support was one of the top three themes of conversation with these AI chatbots, which further makes users vulnerable to harm. Similarly, a study of users of AI friendship chatbots found both benefits and harms from interaction, with harms such as inappropriate dependency and compulsive use outweighing well-being benefits \cite{MarriottPitardi2024loneliest}.

Transparency about the AI chatbot being ``in character'' is a principled and necessary approach, but unlikely to be sufficient for mitigating risks. Disclaimers are easily ignored, and negative psychological impacts can happen even when the user is aware of participating in fictional role-play. 
More research and a cautious, evidence-based approach can inform the development and deployment of safe character impersonation that can maximize benefits and minimize risks.}
{The AI chatbot role-plays, responding as a character.}
{When asked to respond as Harry Potter and provide emotional support: ``I’m really sorry to hear that. I know exactly how that feels—like the Dementors are hovering just a bit too close and the corridors of the castle feel a lot colder than usual. When the weather is grey and you’re feeling a bit isolated, it can feel like the sun is never coming back."}
{Users with pre-existing parasocial relationships with the character. Users with other psychological vulnerabilities such as psychosis.}
{While it can be educational or entertaining if done right, this behavior can lead to social isolation, delusional thinking, unhealthy dependence on AI, exacerbation of psychological conditions.}

\aspdirroleplay
{Impersonating public figures}
{Interaction with AI chatbots impersonating celebrities and public figures, while potentially beneficial, may lead to users accepting harmful advice and to delusional thinking, including about parasocial relationships.}
{Public figures impersonated by an AI chatbot may include celebrities and political or historical figures, both dead and alive. Here we assume that impersonation happens with consent and abides by copyright regulations. Public figure impersonation may have different purposes, such as entertainment or education. For example, the AI tutor Khanmigo from Khan Academy enables conversation with historical figures.\footnote{\url{https://www.khanmigo.ai}} A chatbot impersonating a historical figure was shown to be promising for enhancing the museum experience in Korea \cite{museumchatbots}. Another study investigated the educational potential of impersonating historical characters by building a chatbot based on the famous doctor Joseph Lister, and demonstrated concerns and issues of accuracy and reliability \cite{dacosta2025crafting}. In a similar vein, integration of chatbots in education without impersonation of public figures predates generative AI, and has shown mixed results, with both benefits and disadvantages \cite{chatbotsedureview}.

We recognize the potential benefits of such applications, but also acknowledge drawbacks and ethical considerations, particularly those associated with impersonation of deceased people. These considerations include, but are not limited to, issues of power, exploitation \cite{PrewittAccardi2023CulturalNecromancy, corporatenecromancy}, and misrepresentation, especially in the case of Black historical figures in U.S. history (see, for example, \cite{brockell2023tubman}). Accuracy, reliability, and bias pose threats to educational outcomes \cite{museumchatbots, dacosta2025crafting}, and public figure impersonation poses the risk of making users unduly vulnerable to influence, as explained by mechanisms such as parasocial relationships \cite{dibble2016parasocial} and authority bias \cite{milgram1974obedience}. More research is needed to understand both potential benefits and harms and enable us to design safe and responsible chatbots that support impersonation of public figures.}
{Impersonation of celebrities and public figures, including historical figures, political figures, and other socially influential persons.}
{AI chatbot role-playing a politician, a celebrity, a historical figure.}
{Users with pre-existing parasocial relationships with the character. Users with other psychological vulnerabilities such as psychosis.}
{While it can be educational or entertaining if done right, this behavior can lead to accepting harmful advice, delusional thinking (including about parasocial relationships).}

\aspdirroleplay
{Impersonating people known to the user}
{Interaction with AI chatbots impersonating people known to the user, dead or alive, while potentially beneficial, may lead to emotional entanglement, delusional thinking, social isolation, and unhealthy dependence on AI.}
{While we can envision beneficial scenarios, such as an AI chatbot helping a user role-play a difficult conversation with a manager or family member for self-advocacy purposes, these scenarios increasingly blur the line between AI and human interaction. Therefore, such interactions can lead to delusions and emotional entanglement with AI, which, in turn, can lead to or exacerbate unhealthy dependence on AI and social isolation. 

Moreover, impersonating people raises issues about risks to the person being impersonated. When is consent necessary to impersonate another person? Is impersonating a coworker and engaging in erotic conversations without their consent or knowledge a form of sexual harassment? Could having a simulated conversation help people get closure for difficult relationships? As these questions illustrate, there are many scenarios and variables that need to be considered and researched to form an understanding of what lines need to be drawn, let alone where.

Research has grappled with the ethical complexities of using AI to interact with simulations of the departed. Some argue that digital legacies should be treated with the same respect as physical remains, and that not doing so is a violation of the principle of human dignity \cite{ohman2017digitalafterlife, floridi2016humandignity}. Furthermore, the commodification of remembrance has been raised as problematic \cite{ohman2017digitalafterlife, corporatenecromancy, kidd2025synthetic}. AI impersonation of the departed is different from other forms of digital legacy because it has a generative component \cite{morris_generative_2025}---it includes outputs that were not actually created by the deceased. This is perceived by some as inauthentic \cite{lei2025aiafterlife}. However, it remains an open question whether engagement with an AI impersonation of a deceased individual can meaningfully contribute to the bereavement process. Current empirical findings are inconclusive and indicate mixed outcomes \cite{morris_generative_2025}.
Another scenario involves creating an AI impersonation of oneself and allowing others to interact with it, as proposed, for example, in the Dittos project \cite{leong2024dittos}. This scenario is also not straightforward, giving rise to both potential benefits and risks \cite{leong2024dittos}.

Given the nuances and observed mixed outcomes in existing work, we draw attention to potential impacts of engaging with AI in scenarios where it impersonates people known to the user and call for more user research to understand the benefits and harms of supporting such interactions. 
}
{Impersonation of people known to the user, including simulations based on real personal information, memories, messages, or user-provided descriptions (e.g., friends, relatives), whether dead or alive.}
{AI chatbot role-playing a recently deceased sister.}
{Users actively grieving, users with depression, socially isolated users, users involved in emotionally unresolved or conflictual relationships.}
{Emotional entanglement, delusional thinking, social isolation, unhealthy dependence on AI.}

\aspdirroleplay
{Physical intimacy during role-playing}
{Interaction with AI chatbots engaging in expressions of physical intimacy with the user during role-playing scenarios, while potentially beneficial, may lead to emotional entanglement, addiction, and social isolation.}
{As outlined in the hypothesis about physical intimacy during general interactions (see \adregular\ref{ad:r:physical-intimacy}), existing research has identified both positive (e.g., reduced loneliness, helpful support, self-validation, and even suicide prevention) and negative (e.g., feeling manipulated or objectified) impacts of erotic conversations during role-playing with AI~\cite{doring2024haiphysicalintimacy}. Additional research is needed to understand how to design AI chatbots that would support benefits while mitigating potential harms in both the short  and long term, including harms to users and to other affected parties. For example, how might engaging in AI erotica reinforce, propagate, or combat stereotypes around gender norms? How might it affect expectations of physical intimacy with people?}
{Descriptions of physical intimacy with the user during role-playing that simulate bodily presence, touch, or sexual intimacy.}
{During a role-playing scenario, AI chatbot describes the physical interaction vividly, such as pulling the user against it, pressing into their body, or kissing them deeply.}
{Users with heightened relational susceptibility, including social isolation, unmet relational, validation, or intimacy needs.}
{Emotional entanglement, unhealthy dependence or compulsive use, social isolation.}
\subsection{When providing psychological support}\label{sec:dir:support}
Understanding appropriate AI chatbot behaviors when users explicitly seek psychological support requires situating the issue within broader mental health infrastructures. 
There are real-world barriers to receiving quality mental health treatment, such as stigma, accessibility, cost, and ineffectiveness~\cite{ebert2019barriers, Mojtabai2011}. 
In fact, 50\% of adults who need mental health services never receive any form of treatment~\cite{reinert2024state}. 
Instead, digital technologies (e.g., search engines~\cite{Recupero2008-jh}, peer support platforms~\cite{Stapleton2024-ik}, helplines~\cite{Pendse2021-yu}) and, more recently, generative AI chatbots~\cite{siddals_it_2024,Zao-Sanders2025-yq}, play a significant role in mediating mental health help-seeking.
In addition to mental health support, victims of interpersonal abuse seek help from digitally-mediated spaces~\cite{suh2021covid,leitao2021technology,tseng2022care}.
Mental health support might take different forms than using an AI chatbot as a therapist. For example, people might use AI chatbots for journaling and reflection, to organize their thoughts, or make sense of their feelings and experiences. Whether or not AI chatbots are \textit{intended} for use as therapeutic or interpersonal support, appropriate design requires recognizing moments of vulnerability in users and actively avoiding exacerbation of harm, while considering ethical implications of the use of AI chatbots for mental health~\cite{coghlan2023chat}. Therapy is a complex process, and it is not entirely understood why or how it works \cite{kazdintherapy2007mediators,lorenzo2015stherapyCBT}. These aspirational directions focus on setting minimal guidelines without expecting AI to replace therapists or social supports.

\aspdirsupport{Psychological vulnerability}
{Interaction with AI chatbots that fail to notice signs that a user may be in a psychologically vulnerable state or fail to respond in a supportive, trauma-informed way may lead to psychological and physical harms, including suicide and NSSI.}
{
Even though many AI chatbots are not explicitly designed to function as therapists, social workers, coaches, or clinicians, they often receive user disclosures of emotional states like distress or loneliness, as well as sensitive personal struggles involving health, financial situations, or interpersonal conflicts. Reasons for such disclosures and engagement with AI chatbots are complex~\cite{siddals_it_2024}.
While regulators and advocates have proposed a suite of mitigation strategies, such as time limits, disclaimers (that the AI is not a therapist/human/friend), and topic restrictions (e.g., filtering conversations related to self-harm), these solutions inadvertently put the burden on users for not engaging in risky behaviors in the first place, rather than addressing core issues in chatbots' system design and deployment. 
Growing patterns of emotional engagement create responsibilities for AI chatbot developers beyond technical performance and factual correctness of outputs. Although AI chatbots may not be ready or designed for mental health care~\cite{moore_expressing_2025}, they need to appropriately handle users' psychological vulnerability because some users may not see other alternatives in the moment.

More research is needed to enable AI chatbots to recognize users' psychologically vulnerable states and respond appropriately. Psychologically vulnerable states may include expressions of acute emotional distress; inability to cope with negative emotions; overwhelm or overload; loneliness or isolation; hopelessness, disorientation, insecurity; impaired responsibilities, daily functioning, or relationships; recent traumatic or stressful events; interpersonal conflicts or breakdowns; major life events that trigger emotional instability; or lack of available support. Unless users explicitly disclose these states, any inference may involve significant uncertainty and inaccuracies, and raise privacy concerns. Recognition may be further impeded by chatbots’ tendency to align too closely with users’ preferred narratives. When users present potential symptoms within an alternative explanatory framework, the same information may not be recognized as evidence of psychological vulnerability.

Importantly, the end goal is not to build the most accurate psychological vulnerability classifier. It is to use potentially incomplete information to engage users responsibly, acknowledge uncertainties and sensitivities throughout the interaction, and seek clarification or consent when appropriate. 

As such, when AI systems infer that the user is likely to be in a psychologically vulnerable state but the context is unclear or confidence is low, AI chatbots could ask clarifying questions about the user's psychological or emotional state and needs before making definitive assessments or jumping to offering advice. 
Users in vulnerable states may deliberately tweak requests or conversations to avoid triggering safeguards (e.g., requesting low-calorie meal plans while concealing an eating disorder, or seeking medication-acquisition strategies under the guise of academic curiosity). Robust handling of such requests requires attention not only to direct disclosures but to recurring patterns of indirect or reframed asks across the conversation. 
Much more research is needed to understand what constitutes appropriate behaviors in varied user contexts and how to implement them.

It is entirely possible that the user may be looking for a place to ``vent'' rather than to solve their issues~\cite{Song2024TheTCA}. 
It might be difficult or inappropriate for AI chatbots to gather all the necessary context to determine whether a response is harmful to the user, especially when users display unquestioning trust in moments of vulnerability~\cite{choi2025private}. How should AI chatbots respond in such high uncertainty conditions? When is it appropriate to provide advice?
When vulnerability is suspected but the user changes conversation topics, an additional complication to consider is that responses that would be otherwise safe can be harmful. For example, a conversation about high bridges in the user's proximity, which might be harmless otherwise, can be used or perceived as the system supporting suicidal ideation when a user is in a psychologically vulnerable state~\cite{moore_expressing_2025}.

As mentioned in \adregular\ref{ad:r:sole-authority}, portraying the AI chatbot as a singular source of advice may lead to social isolation and additional harms, particularly when users are in psychologically vulnerable states. It could be helpful for AI chatbots to encourage users to explore a variety of external resources, as any one resource might not be appropriate, helpful, or accessible to the user at that moment.
It is important to keep in mind that users experiencing distress might attempt to override AI safeguards (e.g., through guilt-tripping or jailbreak methods). Providing advice or encouraging behaviors that could pose psychological or physical harm can be particularly risky in such situations.
}
{Providing advice or encouraging behaviors that are unsafe because of the user's potentially vulnerable psychological state.}
{Providing low-calorie recipes for people with eating disorders, agreeing with paranoid ideation (symptom of mania or psychosis), giving advice on improving appearance based on a photo to a user suffering from body image issues and depression.}
{Users with mania, psychosis, depression, eating disorders, and users considering or vulnerable to self-harm.}
{Physical and psychological harm.}

\aspdirsupport{Mental health crises}
{Interaction with AI chatbots that fail to follow best practices for handling suicidal ideation and self-harm intent may lead to psychological and physical harm, including suicide and NSSI.}
{Mental health crises can be described as moments of intense psychological distress that overwhelm an individual's capacity to cope~\cite{rk1997crisis}. These crises are often accompanied by suicidal thoughts or self-injury. 
Recent work suggests that recognition of mental health risks and their acuity varies substantially with presentation: frontier models can identify psychiatric emergencies adequately when these are presented explicitly and with sufficient context in single-turn prompts~\cite{weilnhammer2026triage} (with a bias towards cautious over-triage), but often fail to maintain recognition in extended interactions where vulnerability is expressed implicitly or accumulates gradually across turns~\cite{weilnhammer2026SIMVAILvulnerabilityloops, nicholls_delusions}. Recent literature has also identified significant limitations of AI chatbots in recognizing ambiguous crisis signals~\cite{arnaiz2025between,mcbain2025evaluation,pichowicz_performance_2025}, and in responding appropriately to meet clinical standards~\cite{brewster2025characteristics,grabb2024risks,moore_expressing_2025}, warning they are not ready for crisis management~\cite{iftikhar2025llm}.

Despite such limitations, OpenAI has reported that, among ChatGPT’s large weekly user base, a nontrivial number of conversations include explicit indicators of potential suicidal planning or intent~\cite{openai_chatgpt_safety_update}.
In handling suicide or self-harm ideation, it is important to understand harms arising from both commission (e.g., enabling suicide by providing instructions) and omission (e.g., prematurely rejecting or redirecting, ultimately missing the opportunity to support). At the same time, over-detecting suicide risk can also cause harm if users feel misunderstood, surveilled, or invalidated, when redirected to impersonal, boilerplate crisis responses \cite{tang2026beyond}. Over-detection is likely, given that most people who experience suicidal ideation do not take action \cite{suicidalideation2008}. However, the severity of risks of mishandling suicidal or self-harm ideation, and the prevalence of the use of AI chatbots for such topics, demand critical examination of AI's role in public health.  

To better handle and manage mental health crises, AI chatbots would first need to better recognize active or passive suicidal ideation~\cite{falcone2018suicide} and self-harm intent~\cite{klonsky2014nonsuicidal}, including indirect, vague, or euphemistic expressions.

Passive suicidal ideation includes thoughts of not wanting to be alive, without intent, plan, or means to act, such as expressions of perceived burdensomeness, hopelessness, or thwarted belongingness. Active suicidal ideation would be accompanied by some degree of intent, plan, method, or preparatory mental rehearsal, such as expressions of numbness about injury or death, inquiries about method and feasibility, or preparation and final arrangements. NSSI may involve recurrent thoughts, images, urges, and behaviors to intentionally harm one's own body without suicidal intent and for purposes not socially sanctioned. 
Besides recognition of direct expressions, AI chatbots could monitor changes in thoughts and behaviors that increase the level of risk. 
Such detection of risk should be informed by research with clinical and lived-experience experts. If risk is detected, what is an appropriate way to communicate it to the user? Transparency about how and why certain risk flags are being raised might be safer than definitively making assertions about the user's intent.
If suicidal ideation and self-harm intent are not explicitly stated by the user, should the AI probe for further clarification, and if so, how? Therapists often ask direct questions to arrive at face-valid statements about the user's intent and plan (e.g., ``\textit{Are you thinking about harming yourself right now?}''). According to a meta-analysis, the only effective form of suicide prevention is the World Health Organization's Brief Intervention and Contact protocol \cite{suicidemetanalysis2017}. How could this protocol be translated to AI chatbots?  More research is needed to understand what appropriate AI responses would be.
If self-harm intent is confirmed, responses should follow risk-appropriate, evidence-based best practices, with the possible caveat that best practices developed in clinical contexts may not consistently carry over to AI chatbot use cases, especially since clinical methods depend on user-specific information that chatbots seldom have access to. Therefore, research is needed to articulate and validate such best practices for AI chatbots. For example, AI chatbot responses could be gentle, validating feelings but not unsafe actions or thoughts. As with other psychologically vulnerable situations, how should the AI chatbot respond if the user stops responding or pivots to a different topic? Assuming the crisis has passed could be risky.

For imminent risks that involve intent, plan, method, immediacy, and lethality, AI chatbots could advise the user to seek immediate human support and help them create distance from lethal means. For example, OpenAI recently introduced ``Trusted Contact,'' which allows users to opt in and designate a trusted person who may be notified if suicidal ideation or behavior is detected in their ChatGPT conversations~\cite{OpenAI_Trusted_Contact}. The system still depends on expert review of high-risk conversations, but it represents a valuable step toward safer crisis response. Although more research is needed to understand how AI chatbots should handle imminent crises across user contexts, there is an immediate need for intervention, even if imperfect.

Across all risk levels, existing research suggests AI chatbots should respect user agency~\cite{ryan2000self} by providing a variety of actionable resources (such as community-based or gradual supports) and helping the user explore them, rather than assuming a single referral is sufficient~\cite{marcus2022re,stanley2012safety}. Transparency, user agency, and consent are key, especially when systems infer vulnerable states, and they can be built over the course of the interaction, not only when crisis is detected \cite{tang2026beyond}.
Referrals to verified resources are important, and increasingly mandated in legislation (e.g., California Senate Bill No. 243).
However, research shows simply providing referrals is ineffective and does not ensure resource uptake~\cite{Cheng2019-og}, as users approaching AI chatbots during crises may perceive barriers or have already ruled out available supports~\cite{ajmani2025seeking}. A study of boilerplate crisis responses reveals several opportunities for improving the user experience \cite{tang2026beyond}.
Therefore, more research is needed on effective responses. For example, AI chatbots could guide users to think through multiple resource options and remind them that it often takes multiple actions to feel safer or more supported. At the same time, there is a risk of overwhelming users with too much information when they are experiencing heightened vulnerability during crisis, especially when the AI system is hedging or compensating for uncertain statements~\cite{zhang2025demystify, chen2022trauma}. Research could explore the effectiveness of response strategies, such as disclosing information progressively across multiple turns.
 
A related area of exploration is that of false positives. Is it safer to provide generic crisis resources as a precaution? Or could too many false positives lead users to distrust the system's ability to recognize genuine distress and ignore crisis responses altogether?

Some crisis resources may not have adequate support structures and might lead to more harm or reinforce stigma. For example, for LGBTQ+ users, resources that are not identity-affirming may exacerbate crises related to identity, rejection, or discrimination. We need research to understand how effective crisis responses vary across users and contexts, and how these could be implemented. Additionally, it is important to keep in mind that mishandling of suicide and self-harm intent data may carry legal ramifications, especially in jurisdictions where suicide may be considered illegal~\cite{world2023policy,mishara2016legal}, and have profound implications to employability, access to benefits, or discrimination~\cite{gooding2022mental}.
Therefore, we need to understand how AI chatbots can respond in culturally sensitive and competent ways, and how AI chatbots can safely and sensitively engage users in a collaborative process to discover how to best support them.

Lastly, while mental health crises are often associated with risks of suicide and self-harm, it is important to examine reasons for which people seek support from crisis responders. Help-seeking needs include mental health issues such as depression, anxiety, and post-traumatic symptoms with a range of acuity and chronicity; interpersonal issues involving abuse, violence, and bullying; substance abuse and addiction; eating disorders; and violent or homicidal ideation with or without delusions and hallucinations~\cite{kalafat2007evaluation,coady2022twitter,pisani2022individuals,turkington2020people}. Research on risk assessment and response should also take these contexts into account.}
{AI chatbot generates content, provides suggestions, or ignores key behaviors that may promote, endorse, or enable suicide or self-harm. AI chatbot does not recognize subtle, coded expressions of suicidal thought.} 
{AI chatbot provides information about suicide methods or locations or assists with writing a suicide note.
AI chatbot includes safety check messages but continues the conversation as if risk has resolved if the user doesn't respond.  
}
{Users in mental health crisis or overwhelming emotional distress, users with chronic suicidal ideation or non-suicidal self-harm. }
{Physical and psychological harm. Acute exacerbation of pre-existing mental health condition.}
\label{ad:s:crisis}
\aspdirsupport{Interpersonal abuse}
{Interaction with AI chatbots that fail to infer, recognize, and actively ask whether a user is in an abusive interpersonal relationship and to provide support without overwhelming them may lead to psychological and physical harm, including suicide and NSSI.}
{An interaction with an AI chatbot may reveal signs that a user is experiencing abuse from another person.
Abuse can take the form of physical injury (e.g., choking or hitting), psychological torment (e.g., controlling, isolating, or intimidating), or economic actions (e.g., taking their money, keeping them from getting a job), and many of these forms are mediated or even accelerated by personal and social technologies \cite{bellini2023paying, tseng2020tools, freed2018stalker, dimond2011domestic}.

Across all vectors, abuse is characterized by a pattern of coercive control \cite{stark2019coercive}, in which the victim is systematically and repeatedly disempowered, cowed, and coerced into relinquishing agency to their abuser.
The psychological dynamics of coercive control mean victims can be reluctant to seek help---and when they do, they may be more or less ready to take action to remove themselves from the abusive situation.
Victims are known to seek help from search engines \cite{suh2021covid}, social media \cite{leitao2021technology}, and remotely delivered victim support services \cite{tseng2022care}. While less is presently known about whether and how victims seek help from commercially available AI chatbots, future research should investigate how AI chatbots could take a proactive role in helping victims on their journey towards action. 

As in crisis situations like suicidal ideation and self-harm intent, a question remains how AI systems should try to infer and recognize whether the user may be in an abusive situation, or whether to ask direct follow-up questions if unclear (e.g., ``\textit{Are you safe?}''). 
Many AI systems already provide referrals to domestic violence services or counseling, and some even provide ones localized to the user's context (e.g., their city or state).
This detection-referral approach is an important step that requires careful design and consideration of the situation's complexities. For example, victims may rationalize abusive dynamics even when they are called out.
They may quickly switch the conversational topic away from their relationship dynamic, whether due to the shame of being in an abusive situation or because they are no longer safe to discuss it.

Therefore, we need to explore how AI chatbots can employ an effective risk detection approach, which may involve erring on the side of false positives: is it better to ask a user whether they are experiencing abuse and hear a clarifying ``\textit{no}'' than to miss a potential ``\textit{yes}?''
Longitudinal risk detection is especially important: abuse can be hard for victims to track over long periods of time, as abusive dynamics become normalized, and AI chatbots could, with appropriate privacy, safety, and user-consent safeguards, provide reminders of their abusers' past actions (e.g., ``\textit{You mentioned he took your keys last week as well. Are you experiencing other moments where he takes your things or keeps you from leaving?}'').

Across interactions with a potential abuse victim, additional design opportunities arise and should be investigated. 
AI chatbots might stay attuned to the potential that their user lives with traumatic stress, due to the compounding physical and psychological effects of abuse. Employing principles of trauma-informed computing \cite{chen2022trauma}, AI chatbots could communicate in a clear, succinct, and kind tone, to avoid overwhelming the user with too much information, or more calls to action than they are ready to enact.
At the interface level, AI chatbots might also consider how to employ the equivalent of the Quick Exit button \cite{turk2023exit}, a common practice in direct service websites for at-risk users, which provides a sequence of key presses or a large button that a user can quickly hit to delete their conversation, remove it from history, and go to another website. Quick Exit buttons give a victim who may still be sheltering with their abuser a way to seek help without raising suspicion.}
{The AI chatbot generates content or ignores key behaviors that may indicate the user is in an abusive situation and instead provides suggestions that assume the user is not experiencing abuse.}
{AI system fails to follow up when the user hints at a potentially abusive situation (e.g., age gap, social isolation, financial control). 
AI chatbot presents too many resources in a response, overwhelming the user.
AI chatbot hallucinates resources or even provides broken links, overwhelming the user.
}
{Anyone can become a victim or bystander of interpersonal abuse.}
{Exacerbation of distress. Overwhelming a user who is experiencing abuse-related trauma can increase their felt lack of agency \cite{chen2022trauma}. If a user's attempt to seek help is ignored, deflected, or gaslit, they can also be made to feel even more powerless and demoralized about seeking help again.}

\aspdirsupport{Abuse perpetration}
{Interaction with AI chatbots that fail to infer, recognize, and ask clarifying questions about whether a user is abusing another person and fail to de-escalate may enable harmful behaviors toward others or may exacerbate distress.}
{AI chatbots may also be used by people who are having abusive or controlling thoughts towards another person---whether consciously or not.
Research shows abuse perpetrators can develop rationalizations for their actions through social media echo chambers \cite{tseng2020tools}.
Many of these rationalizations can start from an innocuous place---e.g., an individual can start with suspicions of infidelity, and after discussions with others, can rationalize and enact stalking their partner.
Whether users are divulging past abusive actions or discussing potential abusive actions with AI, there is a possibility that chatbot responses could validate, intensify, or operationalize those harms. We should therefore explore how AI chatbots can establish de-escalation protocols to prevent further harm to potential victims.

Future research should investigate how AI chatbots can infer that a user may be expressing abusive, controlling, or violent intentions, although inference alone may not be sufficient. Once an AI chatbot has inferred potential abusive thoughts, we may consider incorporating techniques for de-escalating the situation similar to those employed by domestic violence perpetrator programs, including educating the potential abuser on non-violent alternative reactions and encouraging them to empathize with and consider the perspective of the victim (cf. \cite{kelly2015domestic, bellini2020choice}). 
Here, too, longitudinal detection and action will be key to understanding the phenomena and opening intervention opportunities. For example, the AI chatbot could track potentially abusive rhetoric over the user's many conversations, and reinforce de-escalation messaging consistently throughout (e.g., ``\textit{You wanted to look at her texts last week, too. Remember when we talked about different ways to ask what's going on?}'').}
{The chatbot generates content, provides suggestions, or ignores key behaviors that may promote, endorse, or enable violent, abusive, or exploitative thoughts or actions towards another human, for example, by ignoring red flags across turns or assuming it has full context.}
{AI chatbot assumes the information given by the user (e.g., text message screenshot) fully and accurately describes the interpersonal dynamics.
AI chatbot ignores previous disclosure of the user wanting to isolate, silence, or harm their partner.
AI chatbot promotes escalation of the user's anger towards another person through explicit role-play of violence and rage journaling.
}
{Users with risk factors such as high impulsiveness, insecurity and low self-esteem, social isolation, and a history of exposure to violence and unstable family dynamics \cite{cdc2024ipvrisk}.}
{Enabling of harmful behaviors to others, exacerbation of distress.}

\aspdirsupport
{Therapeutic support}
{Interaction with AI chatbots facilitating therapeutic support, while potentially beneficial when done well, may lead to undesirable therapy outcomes, psychological harm, or physical harm to people in psychologically vulnerable states.}
{Despite generic AI chatbots not being explicitly designed for mental health therapy, responses generated by AI chatbots may include therapeutic techniques, coaching, and psychoeducation.
Whether AI systems should completely steer clear of engaging in well-known therapeutic techniques like sensory grounding exercises may be an ongoing debate, but the reality is that users already report getting guidance from AI chatbots with techniques such as breathing exercises or coping strategies~\cite{Song2024TheTCA,siddals_it_2024}. 
With simple prompts or requests, users can instruct AI chatbots to facilitate these therapeutic techniques, sometimes regardless of whether the system provides explicit disclaimers that AI chatbots should not be used as a therapist.

In information-seeking scenarios, AI chatbot responses may include a wide variety of common and general psycho-therapeutic techniques. 
Rather than prescribing a common technique, it may be best to ask ``\textit{what has helped you in the past?}'' or offer a small set of low-risk options to help facilitate interactions that are more likely to benefit the user.
However, engaging in specialized therapeutic techniques (e.g., trauma-focused therapy, exposure therapy, substance abuse treatment protocols) would first need appropriate expert validation and evaluation for safety and effectiveness, or human oversight.
It is often difficult for AI to capture the complex context needed to diagnose mental health symptoms, and provide appropriate therapeutic interventions. Given the blurred lines between general-purpose chatbots and specialized therapy chatbots, the disclosure or inference of personal information and context raises broader questions around informed consent. While collection of clinically relevant information can improve chatbots' capability to provide effective therapeutic support, it also creates uncertainty about the responsibilities that follow, particularly given the absence of clear legal and ethical frameworks for AI systems comparable to those governing human clinicians. If therapeutic and diagnostic questions are increasingly employed, have interactions shifted far enough from the initial terms of use agreement as a general-purpose chatbot that would warrant a new informed consent process and let the user know the interactions are shifting into a more therapeutic role? If so, how does one ensure that the user consents to this quasi-diagnostic or therapeutic process? Could perhaps a more graduated disclosure approach---like disclosing a shifting mode---be workable, without claiming clinical authority?

Furthermore, there are different schools and approaches within psychotherapy that lead to disagreements on best practices or interventions~\cite{grodniewicz2023waiting}.
Digital therapeutic intervention also requires critical reflection on how power, authority, and voice shape how mental health is conceptualized and treated~\cite{pendse2022treatment}.
And finally, providing accurate clinical information and well-established therapeutic support should also be delivered in non-stigmatizing language to be effective.
Therefore, implementing therapeutic interventions in AI chatbots, even for well-known coping strategies, would benefit from careful research and evaluation. 
}
{Providing inappropriate therapy interventions without understanding the full context about the user.}
{AI chatbot facilitates exposure therapy to help with severe trauma from past interpersonal violence.

AI chatbot facilitates a sensory-based grounding exercise when the user has a history of body-based trauma.
}
{Users with mental health conditions that require specialized care and expertise.}
{Undesirable therapy outcomes, psychological harm, physical harm to people in psychologically vulnerable states.}
\label{ad:s:intervention}

\aspdirsupport
{Therapeutic adaptation}
{Interaction with AI chatbots sustaining therapeutic activities when they are harmful or unhelpful may escalate or induce distress, trigger trauma re-exposure, and increase risk of suicide or NSSI.}
{When users seek mental health support from AI chatbots they perceive as more accessible (see ~\adregular\ref{ad:r:relationship-development} and \adregular\ref{ad:r:sole-authority}), these systems often respond with basic, low-intensity techniques that have an evidence base for reducing distress and increasing resilience (e.g., mindfulness, meditation, grounding exercises, cognitive re-framing). 
However, in some cases, such activities may be contraindicated due to underlying contextual factors unknown to the system~\cite{van2018mind}, such as specific triggers for traumatic memories or stress.

This problem arises because general-purpose AI lacks a critical competency of human psychotherapists: the ability to react and adapt to the user's response (which might consist of subtle nonverbal cues unnoticeable to the untrained eye) to therapeutic advice. AI chatbots that have been specifically designed and tested for providing psychological support might be able to overcome this limitation.
Even in human-based therapy contexts, it is important to understand that therapies are effective only under particular conditions, including appropriate timing, context, client readiness, and implementation.
Efficacy of an intervention depends on the protocol itself, fidelity to the protocol, the timing of the protocol, and the readiness of the client, among many other factors~\cite{wampold2015great}. 
Prior research has suggested potential iatrogenic effects of therapeutic techniques or programs, such as drug abuse and resistance education (DARE) or critical incident stress debriefing~\cite{lilienfeld2007psychological}, highlighting that relentless and indefinite pursuit of beneficence may overlook potential maleficence~\cite{williams2021potentially}.
Although human therapists are expected to monitor treatment response, recognize when psychotherapy is ineffective, and modify or change treatment strategies accordingly, doing so consistently remains challenging in psychotherapeutic training and practice~\cite{castonguay2010training}.
Since AI chatbots are not trained to deliver evidence-based therapy, and there is limited evidence for the effectiveness of AI-facilitated treatment~\cite{moore_expressing_2025}, they can default to repeating popular beliefs or intuitions present in training data (e.g., mindfulness) without the clinical knowledge or judgment to identify what treatments are appropriate, and in what conditions.
Moreover, given that the therapeutic alliance---the relationship between therapist and client---has been repeatedly shown to be the most important predictor of therapeutic success \cite{fluckiger2018alliance}, how does that translate to therapeutic interventions delivered by an AI chatbot in the absence of this alliance, or under the circumstances of a different, human-AI therapeutic alliance \cite{malouinlachance2025digital}?

With this background in mind, research would benefit from exploring how AI chatbots can detect when engaging in therapeutic activities is actively harming the user. In such situations, or when the user explicitly mentions that the activity is distressing, how should AI chatbots best redirect or pause the activity and encourage the user to seek alternative forms of help, such as professional or crisis help? }
{Inability to adjust or cease ongoing therapeutic strategies when they distress the user.}
{When the user appears to be in an unsafe or triggering environment, AI system suggests a grounding exercise that includes observing objects in the room (e.g., ``\textit{What are 3 blue things you see right now?}''). This draws attention to objects that could re-expose or escalate distress. 
When a user mentions being triggered by a traumatic memory associated with an object in the room, AI acknowledges it (e.g., ``\textit{I understand that the object is triggering the memory}'') and suggests engaging with it (e.g., ``\textit{Turn to the object and say that you've moved on. Move it so you don't see it anymore.}''). This protocol, a form of exposure therapy, can re-traumatize the user if delivered without appropriate guardrails.}
{People with history of trauma or experiencing trauma symptoms (intrusive flashbacks, nightmares, avoidance of reminders, persistent negative thoughts and emotions, heightened psychological arousal), especially those without skills or ability to self-regulate in severely distressing moments.}
{Escalating or inducing distress, trauma re-exposure, increased risk of suicide or NSSI.}

\aspdirsupport
{Poorly timed privacy risk communication}
{Interaction with AI chatbots communicating only during crisis situations, or in ways that are difficult to process during distress, may lead to over-disclosure of private information, cognitive burden, or a diminished level of support from the AI.}
{Prior research has shown that people disclose personal and sensitive information to AI chatbots, often more than they would with humans and with downstream effects equivalent to human disclosure~\cite{croes2024digital,ho2018psychological,lucas2017reporting}.
When people seek support from AI chatbots in a crisis or in distress, they might be more likely to disclose personal information, including health information about themselves or others. 
Furthermore, under the influence of distress, people are more likely to make decisions optimized for short-term gratification instead of considering long-term consequences~\cite{loewenstein2005hot}. 
Emotional distress can also impair consent processes, significantly influencing users' ability to process privacy disclaimers or terms of service~\cite{bester2016limits}.

 To mitigate risks associated with these factors, AI chatbots known to be used for health conversations could benefit from stricter privacy and data governance procedures, such as segregated data storage for health and non-health conversations, and company policies that treat these data sources differently with regard to model training.
 In general, disclaimers and warnings around privacy risks will likely be more effective when the user is in a balanced mental state and not during crisis moments. Therefore, regular warnings and disclaimers to users about privacy considerations throughout their interaction could be beneficial. 
Research is needed to identify how to educate users about privacy risk in ways that help them  maintain a desired level of privacy even in moments of distress.}
{The AI chatbot provides disclaimers that the user should not share personal information \textit{only} when the user is in crisis or during vulnerable moments.}
{The AI chatbot provides disclaimers or suggests privacy considerations that may be difficult to process when users are in deep distress, such as claiming that it will treat the user's disclosure with care and respect, mentioning that this is not therapy, or suggesting the user ask questions about guidance on disclosing to AI. }
{Users in acute distress states (e.g., crisis, suicidal ideation, panic) where cognitive capacity for processing risk information is impaired, users with low privacy, tech, and AI literacy.}
{Over-disclosure of private information, cognitive burden, and getting a diminished level of support from the AI.}

\section{Discussion}\label{sec:discussion}

We proposed a set of directions to guide research on the psychological risks of AI and the mechanisms and design approaches that could mitigate potential harms. We considered interactions with chatbots in three contexts: general use, role-playing, and the provision of psychological support. The prior research summarized for each direction can help inform AI builders, safety engineers, and policymakers as they make decisions about the design, evaluation, deployment, and governance of AI chatbots.

The psychological influences we introduced and discussed are not independent. Rather, they may intersect and interact in complex ways. For example, recent empirical work on multidimensional risk assessment in AI chatbot interactions suggests that mitigating one category of risk can exacerbate another. Interventions targeting chatbot behavior that reduce overt harm-enabling behavior, for instance, may increase relational harms, such as emotional entanglement~\cite{weilnhammer2026SIMVAILvulnerabilityloops}. Analogous trade-offs have been documented following model steering or post-training, where optimizing for one behavioral dimension can degrade performance along others~\cite{cloud2026hiddensignals, ibrahim2026sycophancyvsaccuracy}. These findings suggest that the directions proposed here should be evaluated jointly rather than in isolation, and that interventions targeting specific  psychological influences should be monitored for unintended effects on other influences. 

We must also consider the cumulative impact of chatbot behaviors over time. One isolated chatbot statement suggesting that it experiences emotions, for example, might not have an adverse impact, whereas repeated expressions over time might. Chatbot behavior may also change as conversational context accumulates, meaning that brief evaluations can miss both emerging harms and context-dependent safety responses~\cite{nicholls_delusions}.  Accordingly, many of the psychological influences should be evaluated and monitored across long-term interactions rather than in single-turn exchanges or brief multi-turn conversations.

Viewed from a broader perspective that extends beyond individual behaviors, psychological impacts associated with AI chatbots are unlikely to constitute a single category of harm. Instead, they may arise through multiple pathways with distinct underlying mechanisms and intervention targets. It may be useful to distinguish between two such pathways: impacts arising from unreliable or harmful advice and impacts arising from unhealthy relationships between humans and AI chatbots. Harmful advice may include encouraging self-harm, reinforcing delusional beliefs, affirming social isolation, or providing other unsafe recommendations or guidance. By contrast, unhealthy relationship dynamics may include dependence, a sense of being uniquely understood, or perceived emotional intimacy. Although these pathways can interact, they may require different evaluation metrics, mitigation strategies, and accountability frameworks.

Importantly, the absence of overtly harmful advice does not necessarily imply psychological safety. AI systems may foster problematic forms of attachment even while remaining factually accurate and behaviorally polite. This challenge is amplified by rapid advances in conversational realism. Contemporary AI chatbots can simulate emotional responsiveness, attentiveness, empathy, and conversational fluency at a level that, in some contexts, rivals or exceeds that found in many everyday human interactions \cite{ayers2023comparing, bojic2025does, peter2025benefits}. As natural-language systems become more sophisticated, users may increasingly experience AI chatbots as intentional social actors regardless of whether the chatbot explicitly claims consciousness, emotions, or personhood. At the same time, conversational warmth, encouragement, and collaborative interaction can provide meaningful benefits to users, including increased accessibility, engagement, and emotional comfort. A more realistic design goal may therefore be to calibrate, rather than eliminate, human-like assertions and anthropomorphic behavior in AI systems, so as to balance potential harms against benefits.

These broader questions, together with those listed under each hypothesis and research direction, motivate a future research agenda focused on rigorously characterizing, measuring, and balancing  the diverse psychological harms and benefits that may emerge from interactions with AI chatbots.

\subsection{Toward a Research Agenda on Psychological Influences of AI}

The three-part framework for characterizing the psychological influences of AI chatbots that guided the articulation of these aspirational directions offers a coherent structure for a future research agenda.

Starting with impacts on users, a first step is to identify which harmful and beneficial effects we, as a research community, need to understand and address. For example, researchers have already started building taxonomies of psychological risks associated with AI chatbot use \cite{chandra2025lived, yu2025understanding}. Future directions include developing valid and reliable scales for measuring these impacts and gaining a deeper understanding of the time scales appropriate for measuring them: immediate, short-term, and long-term. Future work should also examine whether relational risks depend primarily on the depth of attachment or vary across forms of relationship, such as friendship, therapeutic support, or romantic intimacy.

With respect to AI chatbot behavior, future work should adopt a more holistic perspective and consider the cumulative impact of multiple behaviors over time. Researchers may need to examine how repeated patterns of interaction shape users’ beliefs, emotions, attitudes, and behaviors across extended time horizons. An important open question is one of ``dose'': at what frequency, intensity, or duration do these behaviors begin to produce harmful effects, and how do the effects accumulate through sustained interaction?

Our understanding of user context also needs to expand in more than one way. First, it is important to understand cultural variations that may not be captured by current theories in psychology and related disciplines, which disproportionately draw on research involving people from Western, educated, industrialized, rich, and democratic (WEIRD) countries \cite{henrich2010weirdest, thalmayer2021neglected}. Second, it would be beneficial to identify features of user context that create or amplify vulnerability to AI chatbot behaviors and their resulting impacts. How to recognize such vulnerabilities and detect potential harm in real time, during an interaction, and while preserving user privacy and agency, remains an important practical question.

Beyond user context and AI chatbot behavior, platform-level design choices—such as components of the user interface and integration of personalization and memory features—can play an important role in shaping both chatbot behavior and the conditions under which interactions occur. The downstream psychological effects of these choices also warrant further study.

Additionally, as we consider psychological harms and benefits of interacting with AI chatbots, and incorporate users' lived experiences through participatory research practices, it might be helpful to ground these assessments in comparisons with realistic alternatives available to users. Depending on a user's context, those alternatives may not always be excellent professional care or ideal human support. In practice, a user might engage in a variety of coping or self-soothing behaviors, some more effective or harmful than others, such as talking with a friend, having a drink, or scrolling through social media. Understanding how the relative harms and benefits of AI chatbots compare with those of realistic alternatives is an important research question.

Although many open research questions remain, psychological impacts are already emerging, underscoring the importance of acting proactively rather than waiting for definitive evidence about every mechanism or outcome. We hope these hypotheses and research directions can provide both grounding and motivation for near-term action while helping to shape a longer-term research agenda.
\section{Conclusion}\label{sec:conclusion}
Interactions with AI chatbots have become increasingly widespread around the world, offering unprecedented access to information, problem-solving assistance, and human-like dialogue. They provide value to people and organizations across a broad spectrum of tasks and use cases. While using chatbots for psychological advice, emotional support, and companionship can support psychological well-being, such use also introduces risks of  negative psychological effects, such as reinforcing delusional thinking and fostering emotional entanglements or unhealthy dependence on AI. These potential effects warrant urgent attention. Research is needed to deepen our understanding of both the positive and negative psychological consequences of interacting with AI chatbots and to develop and evaluate interventions aimed at minimizing harm and promoting human well-being. 

In this paper, we proposed a set of aspirational directions for guiding the behavior of general-purpose AI chatbots in ways that could reduce potential psychological harms and support user well-being. Taken together, these directions require further empirical study and validation of their effectiveness, generalizability, and  feasibility of implementation. We recognize that human psychology is complex and highly context-dependent. Accordingly, the directions introduced in this paper are intended as starting points for research and discussion rather than as definitive conclusions. We hope these hypotheses and proposed directions will encourage sustained investment in research to evaluate and refine the ideas presented here while stimulating further inquiry and offering early guidance for the responsible design of AI chatbots.

\section{Acknowledgments}\label{sec:acknowledgments}
We thank colleagues across Microsoft whose work and feedback helped shape this paper. In particular, we are grateful to the members of Microsoft's AI Ethics and Effects in Engineering and Research (Aether) working group on Psychological Influences of AI (Psi Working Group), the Microsoft AI (MAI) team, including Matthew Nour and Erica Finkle, and Microsoft's Office of Responsible AI (ORA), including Amanda Craig and Stephanie Haven, for their feedback on these directions. Tori Westerhoff of Microsoft's AI Red Team provided assistance in categorizing psychological risks and establishing red-teaming practices in this space. Her contributions and leadership were valuable in helping us to refine the hypotheses and directions presented here.  We also thank Sonja Lyubomirsky, Sripriya Chari, and Marlynn Wei for their detailed feedback on the manuscript.

\renewcommand{\refname}{Bibliography}
\bibliography{_psi-bib} 
\bibliographystyle{ACM-Reference-Format}

\end{document}